\documentclass[12pt]{article} 

\usepackage{srcltx} 
\usepackage{amsmath,amsxtra,amssymb,amsthm}
\usepackage{times,latexsym,array,verbatim,url}
\usepackage{color,graphicx,cite}
\usepackage{datetime}
\usepackage{booktabs}
\usepackage{caption}
\usepackage{multirow}
\usepackage[linesnumbered,ruled,vlined]{algorithm2e}
\usepackage{graphicx}
\usepackage{graphics}
\usepackage{orcidlink}

\newtheorem*{tips*}{Suggestion}
\newtheorem*{temp*}{Template}

\theoremstyle{remark}

\title{\LARGE \bf HCVR: A Hybrid Approach with Correlation-aware Voting Rules for Feature Selection}

\date{}
\author{Nikita Bhedasgaonkar\footnote{Nikita Bhedasgaonkar an undergraduate student at AI and Data Science Department at PICT, Pune, who is also pursuing online B.S. in Data Science and Applications at IIT Madras, was a research intern at IIT Bombay where this work was carried out.},  Rushikesh K. Joshi \textsuperscript{\orcidlink{0000-0002-2712-1406}}
 \\
{\normalsize Department of Computer Science \& Engineering }\\ {\normalsize  Indian Institute of Technology Bombay} \\ {\normalsize Powai, Mumbai - 400076, India.}\\ 
{\small \em Email:nbhedasgaonkar@acm.org, rkj@cse.iitb.ac.in} }

\begin{document}

\maketitle
\thispagestyle{empty}
\pagestyle{empty}

\begin{abstract}
 In this paper, we propose HCVR (Hybrid approach with Correlation-aware Voting Rules), a lightweight rule-based feature selection method that combines Parameter-to-Parameter (P2P) and Parameter-to-Target (P2T) correlations to eliminate redundant features and retain relevant ones. This method is a hybrid of non-iterative and iterative filtering approaches for dimensionality reduction.  
 It is a greedy method, which works by backward elimination, eliminating possibly multiple features at every step. The rules contribute to voting for features, and a decision to keep or discard is made by majority voting. The rules make use of correlation thresholds between every pair of features, and between features and the target. We provide the results from the application of HCVR to the SPAMBASE dataset. The results showed improvement performance as compared to  traditional non-iterative (CFS, mRMR and MI) and iterative  (RFE, SFS and Genetic Algorithm) techniques. The effectiveness was assessed based on the performance of different classifiers after applying filtering.
\end{abstract}

\section{Introduction} \label{sec:intro}

Feature Selection (FS) is a critical preprocessing step in any machine learning (ML) pipeline, which aims to identify the most informative and relevant features from a high dimensional dataset.~\cite{guyon2003introduction}. By eliminating redundant or irrelevant features, FS may be used to improve model performance, accuracy and to reduce computational overhead during training and inference~\cite{dy2004feature}. FS is particularly important in deep learning(DL) approaches where the models are often overparameterized and prone to overfitting especially when working with limited labeled data or noisy inputs.

In domains such as software engineering, bioinformatics and cybersecurity, datasets tend to be high-dimensional and contain complex inter-feature correlations. In such scenarios, even powerful models such as deep neural networks struggle to isolate semantically relevant features~\cite{yang2022rethinking}. While DL models are theoretically capable of learning intricate non-linear relationships, they may get overloaded because of redundant and irrelevant features in order to reduce their influence. Therefore, FS is helpful for sensible dimensionality reduction.

Traditional FS methods are broadly classified into filter, wrapper, embedded, and hybrid approaches~\cite{guyon2003introduction, dy2004feature}. Examples of filter-based methods include Mutual Information (MI)~\cite{bennasar2015mi}, Correlation-Based Feature Selection (CFS)~\cite{hall1999cfs}, and Minimum Redundancy Maximum Relevance (mRMR)~\cite{peng2005mrmr}. Examples of wrapper-based techniques include Recursive Feature Elimination (RFE)~\cite{scikitrfe}, Sequential Forward Selection (SFS)~\cite{scikitsfs}, and a Genetic Algorithm (GA)~\cite{deb1999intro}. For our context, we classify them into {\em iterative} filters and {\em non-iterative} filters.

\subsection{Brief overview of our approach}
In this work, we propose HCVR (Hybrid Approach with Correlation-aware Voting Rules for Feature Selection), a novel hybrid FS algorithm, which combines non-iterative and iterative filters for dimensionality reduction. The HCVR approach incorporates the following key perspectives:
\begin{itemize}
    \item \textbf{Parameter-to-Parameter (P2P) and Parameter-to-Target Correlation (P2T):} which respective capture pairwise correlation among the features, and correlation between features and the target. 
    \item \textbf{Rules based on P2P and P2T values:} They help us to identify which features to prioritize and which to de-emphasize. These choices are made based on correlation thresholds, which are learnt through iterative backward elimination.  
    \item \textbf{Majority Voting:} A feature is selected if majority voting is in its favor. Each triple involving the features f1, f2 and the target T gives us one correlation value for $f1$-$f2$, and two correlation values for $f1$-$T$ and $f2$-$T$. Each triplet of these correlation values contributes to one vote count for $f1$ and one vote count for $f2$. A vote count is either 0 or 1.
    The final decision for a feature is made based on majority aggregate vote counts for it.
\end{itemize}

To evaluate the effectiveness of our proposed approach, we evaluate it on the SPAMBASE dataset from the UCI Machine Learning Repository~\cite{spambase1999}. This dataset is a widely recognized benchmark for binary classification tasks. Our method is rigorously compared against both non iterative filter-based and iterative filter-based FS techniques. 

 Unlike existing correlation-aware approaches, HCVR does not use objective functions, but relies on formulation of logically sensible rules to vote on features. Experimental results on the SPAMBASE dataset demonstrate that our method performed better classification than most cases which we compared with, with the exception of a few cases, in which it produces reasonably at-par results. 

 \subsection{Organization}

In the next section, we primarily provide a brief review covering FS methods with which we compare our HCVR approach. Next,
the HCVR approach is outlined. Finally the results obtained on the SPAMBASE dataset are discussed.

\section{Literature Review} \label{sec:litreview}

Feature selection methods can be broadly classified into non-iterative filters and iterative filters. 

Filters can be applied at once (apriori) before the training algorithm, or iteratively during training. Filters that work apriori, use metrics over
the data itself to decide on importance of features to the target. These typically include correlation based metrics. Filters that decide which features to keep and which to drop iteratively make use of performance of the training algorithm, which may include metrics such as the final accuracy achieved by the algorithm, or the feature weights assigned by the the algorithm. Iterative methods may consider features one by one, either to add or to remove them.
They may also consider different subsets in every iteration. Iterative methods can be found in wrapper~\cite{kohavi1997wrappers} and embedded~\cite{yang2022rethinking} FS methods.
We use three non-iterative methods,  and three iterative methods to compare the performance of HCVR. The non-iterative filtering methods considered are
correlation based feature selection (CFS) ~\cite{hall1999cfs},  Mutual Information (MI)  \cite{bennasar2015mi} and mRMR (Maximum Relevance and Minimum Redundancy)~\cite{peng2005mrmr}, while the iterative methods considered are Recursive Feature Elimination (RFE)~\cite{scikitrfe}, SFS (Sequential Feature Selection)\cite{scikitsfs} and  Genetic Algorithms (GA)~\cite{deb1999intro}.

The HCVR approach is a hybrid approach which aims to combine the merits of non-iterative and iterative approaches. The filter works only on properties of the data like the non-iterative filter approaches, but features are eliminated in backward direction starting from the full set in every iteration by incrementing a threshold value used for feature selection.

\section{Methodology} \label{sec:methodology}

In this section, we first introduce the notations and then develop the core contribution of the method, which is a table of voting rules for reasoning about redundancy and relevance of features, and a voting mechanisms based on votes obtained from the rules.

\subsection{Notations and Definitions.}
Let \( \mathcal{F} = \{f_0, f_1, \ldots, f_{n-1}\} \) be the set of features and \( T \) the target variable. For any pair \( (f_i, f_j) \in \mathcal{F} \), we define:
\begin{itemize}
  \item \( \rho(f_i, f_j) \): Correlation between features \( f_i \) and \( f_j \),
  \item \( \rho(f_i, T) \): Correlation between feature \( f_i \) and the target \( T \),
  \item \( \rho(f_j, T) \): Correlation between the feature \( f_j \) and the target \( T \).
\end{itemize}

A threshold \( \theta \) (e.g., 0.2) is defined during training to classify correlation as:
\[
\text{High (H)} \iff |\rho| \geq \theta; \quad \text{Low (L)} \iff |\rho| < \theta
\]

Threshold tuning is a crucial step in our method, where features are selected or eliminated based on which side of the threshold their Pearson correlation values lie. The threshold determines the cut-off point beyond which features are considered to be dropped, since they are likely to be either redundant or irrelevant. Setting an appropriate correlation threshold is essential to strike a balance between
redundancy and relevance. This threshold needs to be tuned empirically through validation experiments, as it can significantly influence the quality of the selected feature subset and consequently the overall model performance. Here, the hybrid approach achieves the tuning of the threshold via iteration, whereas, the non-iterative part in the method computes the feature correlation scores for $P2P$ and $P2T$ correlations.

\subsection{Voting Rules}

\begin{table}[t]
\centering
\caption{Correlation-aware Pairwise Voting Rules}
\label{tab:decision-rules}
\begin{tabular}{|c|c|c|c|c|}
\hline
\( \rho(f_1, f_2) \)  & \( \rho(f_1, T) \)  & \( \rho(f_2, T) \)   & Vote to f1  & Vote to f2 \\
\hline
H & L & L & 0  & 0 \\
H & H & L & 1  & 0 \\
H & L & H & 0  & 1\\
H & H & H & $ \rho(f_1, T) >= \rho(f_2, T) $ & $ \rho(f_1, T) < \rho(f_2, T) $   \\
L & H & L & 1  & 0 \\
L & L & H & 0 & 1\\
L & H & H & 1 & 1 \\
L & L & L & 0 & 0 \\
\hline
\end{tabular}
\end{table}

The decision to retain or discard features depends on their correlation patterns. Table~\ref{tab:decision-rules} summarizes the rule set used to award voting scores (votes) to each feature based on P2P and P2T correlations.
The votes are collected for every pairing. So, for $n$ features in the feature set, each feature is voted $n-1$ times considering its every possible pairing.
Each vote considers redundancy within the pair, and relevance of each of them to the target. 
Each feature is vetted based on how frequently it is marked for retention or elimination across all pairwise votes. Features with higher keep counts than discard counts are selected for the final subset.

\begin{figure}[t]
  \centering
  \includegraphics[width=0.9\linewidth]{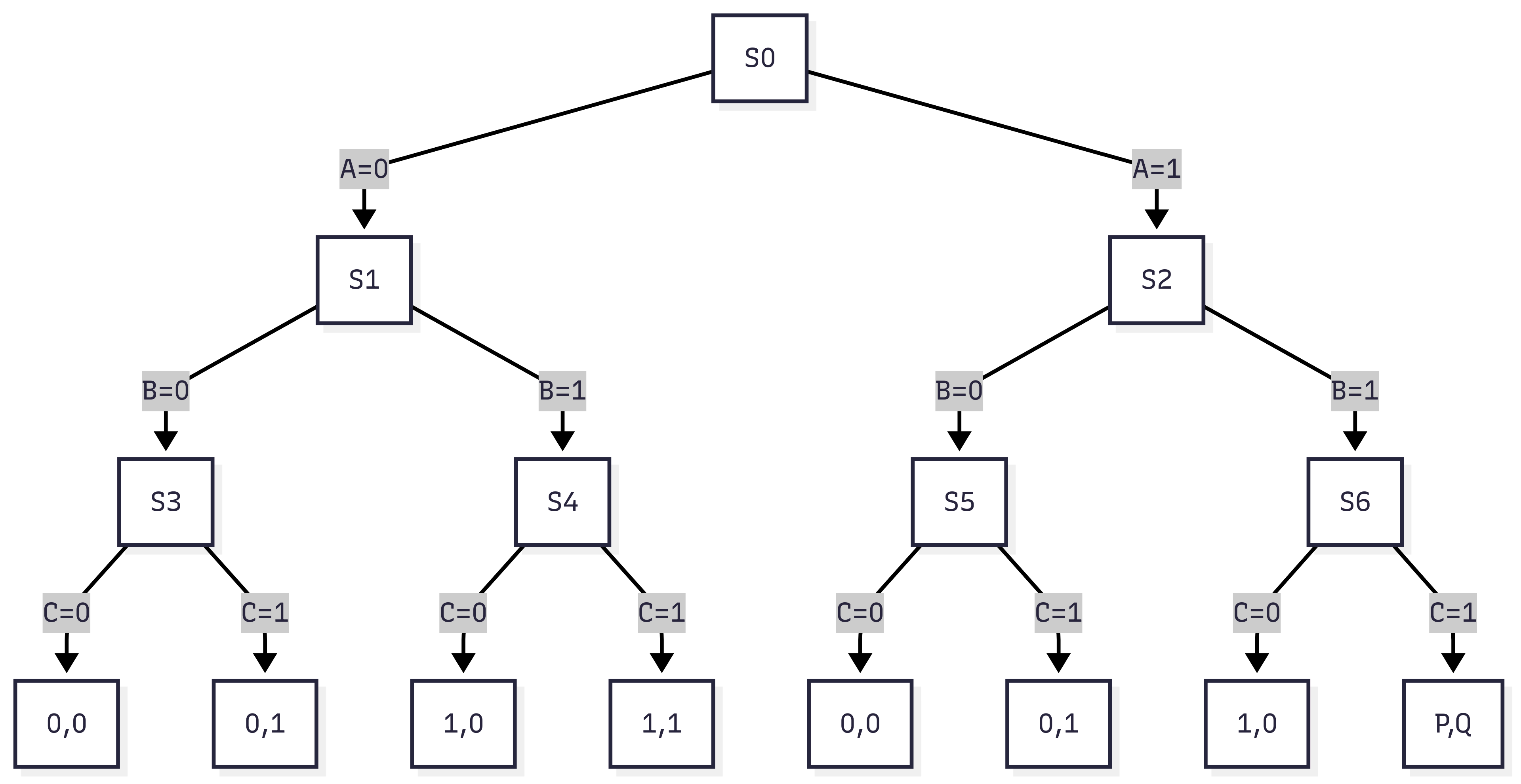}
 
\vspace{1em}

\noindent
\textbf{Symbol Encodings:}
\begin{align*}
\textbf{A} : \rho(f_1, f_2), \hspace{1cm}
\textbf{B}  : \rho(f_1, T), \hspace{1cm}
\textbf{C} : \rho(f_2, T) \\
\textbf{P} : \rho(f_1, T) >= \rho(f_2, T), \hspace{1cm}
\textbf{Q} : \rho(f_1, T) < \rho(f_2, T)
\end{align*}
 \caption{Iterative Refinement of Threshold  with Coarser Backward Elimination}
  
  \label{fig:rule-tree}
\end{figure}

Figure \ref{fig:rule-tree} represents the set of rules of the HCVR model as a hierarchical decision tree. It makes sequential decisions based on correlation-based conditions. Each internal node splits the input based on a binary decisions, and each path from root to leaf represents a distinct rule leading to a final outcome such as \texttt{00}, \texttt{01}, \texttt{10}, or \texttt{PQ}.  At the leaf, the two binary values capture the votes respectively for the two features. Values $P$ and $Q$ encode a conditional vote as shown in the figure.

\section{Experimental Setup}

To demonstrate the performance of the proposed HCVR algorithm, we utilize the \textbf{SPAMBASE} dataset from the \textit{UCI Machine Learning Repository}~\cite{spambase1999}. The SPAMBASE dataset is a widely-used benchmark for binary classification tasks, designed to differentiate between spam. The dataset includes 4,601 instances with 57 numeric features. The target labels are binary values, \texttt{1} marking SPAM or \texttt{0}  marking a NON-SPAM. The dataset does not contain missing or unlabeled values. An 80-20 split is applied to respectively obtain the training and the testing sets. 

The experiments were conducted using Google Colab environment. The experiments did not require a GPU. The programming environment consisted of Python 3.10 and Scikit-learn library~\cite{scikituserguide} for the machine learning classifiers that were used.  The HCVR employs a hybrid approach with a purely correlation-aware voting rules for feature selection. The threshold used inside HCVR is tuned in a hybrid approach by applying HCVR method on the chosen classifier.

\section{Results on the SPAMBASE Dataset}

\begin{table}[t]
\centering
\caption{Comparing Accuracy of HCVR with Filters and Wrappers}
\label{tab:accuracy_comparison}
\begin{tabular}{l|ccc|ccc|c}
\toprule
& \multicolumn{3}{c|}{Filters (Non-iterative)}&\multicolumn{3}{c|}{Wrappers (Iterative)} & Hybrid  \\
\textbf{Classifier} & \textbf{CFS} & \textbf{mRMR} & \textbf{MI} & \textbf{RFE} & \textbf{SFS} & \textbf{GA} & \textbf{HCVR} \\
\midrule
Random Forest   & 90.22 & 91.09 & 92.07 & 90.11 & 90.22 & 94.02 & 93.91 \\
Decision Tree   & 87.61 & 89.24 & 87.50 & 85.98 & 88.26 & 90.22 & 93.15 \\
SGD             & 86.30 & 57.72 & 86.74 & 86.52 & 87.17 & 89.78 & 85.53 \\
SVM             & 82.61 & 68.48 & 98.07 & 88.48 & 88.59 & 89.67 & 74.46 \\
MLP             & 90.33 & 89.24 & 90.43 & 89.78 & 89.02 & 92.83 & 91.74 \\
\bottomrule
\end{tabular}
\end{table}

\begin{table}[t]
\centering
\caption{Comparing Precision of HCVR with Filters and Wrappers}
\label{tab:precision_comparison}
\begin{tabular}{l|ccc|ccc|c}
\toprule
& \multicolumn{3}{c|}{Filters (non-iterative)}&\multicolumn{3}{c|}{Wrappers (iterative)} & Hybrid  \\
\textbf{Classifier} & \textbf{CFS} & \textbf{mRMR} & \textbf{MI} & \textbf{RFE} & \textbf{SFS} & \textbf{GA} & \textbf{HCVR} \\
\midrule
Random Forest   & 90.32 & 92.31 & 94.40 & 90.26 & 90.40 & 94.03 & 95.63 \\
Decision Tree   & 87.50 & 88.80 & 87.06 & 86.03 & 88.31 & 90.21 & 93.83 \\
SGD             & 88.60 & 100.00 & 91.88 & 86.77 & 87.34 & 89.87 & 83.85 \\
SVM             & 83.90 & 66.89 & 93.97 & 88.70 & 88.70 & 89.72 & 71.35 \\
MLP             & 92.16 & 90.74 & 92.18 & 89.98 & 89.17 & 92.87 & 90.89 \\
\bottomrule
\end{tabular}
\end{table}

The comparative analysis of accuracy and precious are shown in Tables \ref{tab:accuracy_comparison} and \ref{tab:precision_comparison} respectively. The tables highlight the performance of the HCVR technique compared to three iterative and three non-iterative filtering techniques when applied over the SPAMBASE dataset.  In the HCVR approach, the jump step for threshold increment was set to 0.02.  It can be noted that in the Tables, the three non-iterative filter methods (CFS, mRMR and MI) were assigned to select the 10 best features and then the classifiers were applied on that set of features, and the wrapper methods were given the full feature set. The three wrappers RFE, SFS and GA selected 5, 5 and 34 features respectively across all classifiers. As shown in Table \ref{tab:accuracy_comparison}, HCVR achieved better or comparable accuracy barring a couple of exceptions. 
 Notably, it performed better than all iterative and non-iterative filters with Decision Tree, and most filters with MLP and Random Forest. It performed
 better than one non-iterative filter for SGD and SVM. HCVR did not perform better than SVM with GA wrapper.
   Similar results were obtained for precision as shown in Table \ref{tab:precision_comparison}.

\begin{table}[t]
\centering
\caption{Comparing Accuracy with best Results of Non-iterative Methods}
\label{tab:accuracy_comparison_k}
\begin{tabular}{lcccc}
\toprule
\textbf{Classifier} & \textbf{CFS} & \textbf{mRMR} & \textbf{Mutual Info} & \textbf{HCVR} \\
\midrule
Random Forest & 93.91 (55) & 93.91 (43) & 93.80 (55) & 93.91 (T=0.04, 54) \\
Decision Tree & 92.26 (54) & 93.04 (43) & 92.93 (54) & 93.15(T=0.02, 45) \\
SGD           & 91.31 (51) & 80.21 (43) & 91.08 (38) & 85.53(T=0.26, 7) \\
SVM           & 92.17 (53) & 74.45 (1) & 92.17 (49) & 74.46(T=0.34, 7) \\
MLP          & 93.80 (51) & 91.40 (52) & 94.02 (54) & 91.74(T=0.08, 47) \\
\bottomrule
\end{tabular}
\end{table}

Next, Tables \ref{tab:accuracy_comparison_k} and \ref{tab:precision_comparison_k} compare the same performance of HCVR with the best found feature sets in the non-iterative filters (CFS, mRMR, Mutual information), which were found by iterating over them by varying the number of features step by step by brute force. For the results found in these tables, MI and CFS use $k$-best strategy, where the top-$k$ features are selected based on relevance scores computed via ANOVA F-statistics (f\_classif) for CFS, and mutual information (mutual\_info\_classif) for MI. For both methods, $k$ is varied over the entire feature range, and the configuration yielding the highest classification accuracy is chosen as optimal found.  The 
Minimum Redundancy Maximum Relevance 
mRMR algorithm  finds out given $k$ the best $k$ featres.
by ranking features from the entire feature set, by assessing their relevance to the target and redundancy with other features. 
We iterated mRMR to find out the best $k$ for it.

\begin{table}[t]
\centering
\caption{Comparing Precision with best Results of Non-iterative Methods}
\label{tab:precision_comparison_k}
\begin{tabular}{lcccc}
\toprule
\textbf{Classifier} & \textbf{CFS} & \textbf{mRMR} & \textbf{Mutual Info} & \textbf{HCVR} \\
\midrule
Random Forest & 95.63 & 93.91 & 95.12 & 95.63 \\
Decision Tree & 93.26 & 93.04 & 94.03 & 93.83 \\
SGD           & 91.30 & 78.26 & 92.07 & 83.85 \\
SVM           & 92.17 & 71.34 & 94.66 & 71.35 \\
MLP           & 93.80 & 91.24 & 94.66 & 90.89 \\
\bottomrule
\end{tabular}
\end{table}

The HCVR approach can be seen to perform better or comparably with these best found results.
In this experiment, the number of selected features in each method is given in brackets in each cell. The HCVR values also show the best threshold found for each classifier along with the number of features that the thresholds select. Due to its hybrid nature, the HCVR approach chooses the corresponding best threshold for each classifier. 

%These experimental results highlight the effectiveness of the voting rule based HCVR dimensionality reduction approach. 

\begin{figure}[t]
  \centering
  \includegraphics[scale = 0.45]{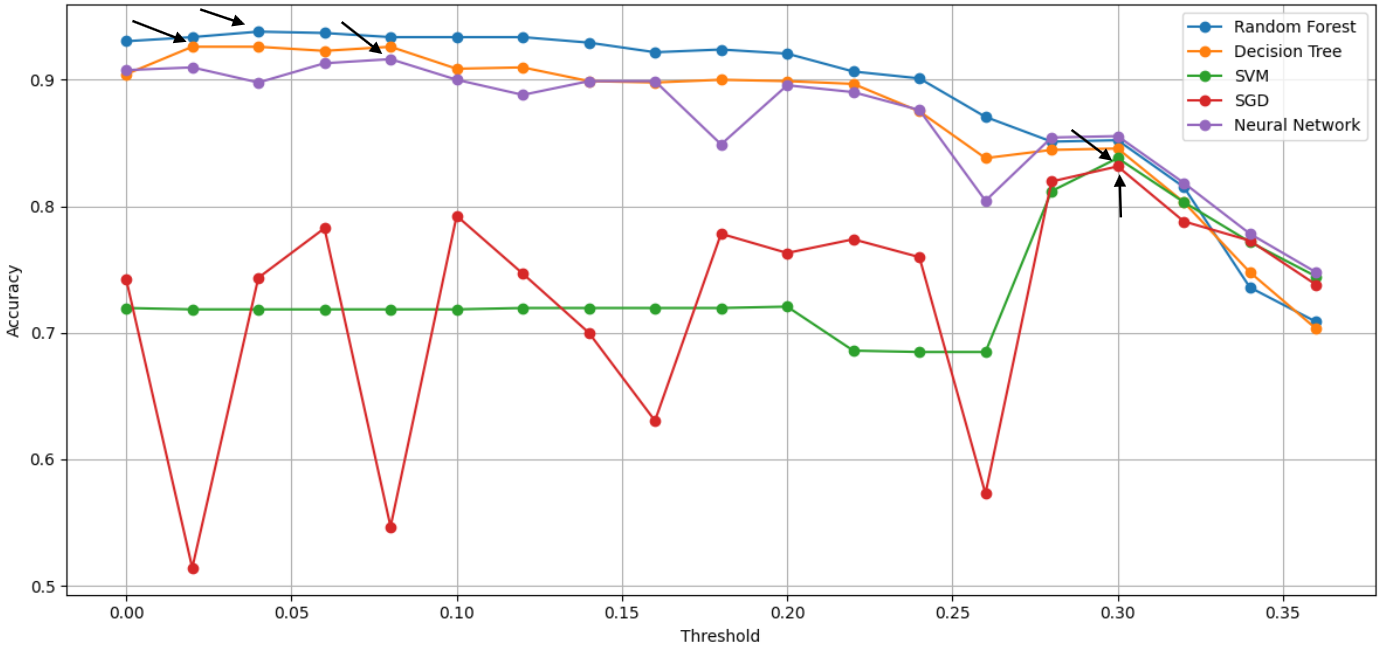}
  \caption{Iterative Refinement of Threshold  with Finer Backward limination }
  \label{fig:accuracy_fine}
\end{figure}

Figure \ref{fig:accuracy_fine} captures the iterative refinements carried out for finding threshold values for use in the rules. The iterative refinements in HCVR threshold for each of the five classifiers are shown. The best values (shown by arrow marks in the figure) found from the training set are carried forward into testing. %As per  the thresholds, 54 features were selected for Random forest, 45 for Decision Tree, 7 for SVM, 7 for SGD and 47 for MLP. 
As the thresholds increase further, the accuracy drops considerably since relevant features get eliminated due to significant drop in the  threshold values.

\section{Conclusions and future work} \label{sec:conclusion}

We introduced HCVR, a hybrid rule-based feature selection approach 
with its correlation-aware voting rules. The approach employs multi-feature backward elimination guided by votes obtained from rules that make use of P2P and P2T correlation scores. A correlation threshold is used to vote on features to remove redundant or weakly relevant features, and also features that are unlikely to impact the target. The threshold are found in the  training phase. Each classifier may deduce with a different threshold. Experimental results on the SPAMBASE dataset demonstrate that HCVR achieves superior or competitive performance across multiple classifiers, as compared to both non-iterative filtering techniques (CFS, mRMR, MI) and iterative filtering algorithms (RFE, SFS, GA). Simplicity and interpretability are two major properties of HCVR. As future work, the technique can be broadened by exploring alternative thresholding methods, alternative voting methods, and by further exploring the interplay between the rules.

% \bibliographystyle{IEEEtran}
% \bibliography{papertemplate}


\end{document}